\documentclass[runningheads]{llncs}
\usepackage{graphicx}
\usepackage{amsmath,amssymb} % define this before the line numbering.
\usepackage{color}
\usepackage[table,x11names]{xcolor}
\usepackage{xcolor,colortbl}
\newcolumntype{C}[1]{>{\centering\let\newline\\\arraybackslash\hspace{0pt}}m{#1}}

\definecolor{Gray}{gray}{0.85}
\definecolor{LightCyan}{rgb}{0.88,1,1}

\newcolumntype{a}{>{\columncolor{Gray}}c}
\usepackage[width=122mm,left=12mm,paperwidth=146mm,height=193mm,top=12mm,paperheight=217mm]{geometry}

\begin{document}

\pagestyle{headings}
\mainmatter

\title{Effective Use of Synthetic Data for\\Urban Scene Semantic Segmentation\thanks{This work was supported by the Australian Centre of Excellence for Robotic Vision.}} 

\titlerunning{Effective use of synthetic data for urban scene semantic segmentation}

\authorrunning{F. Saleh, S. Aliakbarian, M. Salzmann, L. Petersson, J. Alvarez}

\author{Fatemeh Sadat Saleh$^{1,2}$\orcidID{0000-0002-3695-9876​}, Mohammad Sadegh Aliakbarian$^{1,2,3}$\orcidID{0000-0003-3948-6418}, Mathieu Salzmann$^{4}$\orcidID{0000-0002-8347-8637}, Lars Petersson$^{2}$\orcidID{0000-0002-0103-1904​}, and Jose M. Alvarez$^{5}$\orcidID{0000-0002-7535-6322}}
%\index{Saleh, Fatemeh Sadat}
%\index{Aliakbarian, Mohammad Sadegh}
%index{Alvarez, Jose Manuel}

\institute{$^{1}$ ANU, $^{2}$ Data61-CSIRO, $^{3}$ ACRV, $^{4}$ CVLab, EPFL, $^{5}$ NVIDIA\\
	\email{ \{fname.lname\}@data61.csiro.au, mathieu.salzmann@epfl.ch, josea@nvidia.com}
}

\maketitle

\begin{abstract}
Training a deep network to perform semantic segmentation requires large amounts of labeled data. To alleviate the manual effort of annotating real images, researchers have investigated the use of synthetic data, which can be labeled automatically. Unfortunately, a network trained on synthetic data performs relatively poorly on real images. While this can be addressed by domain adaptation, existing methods all require having access to real images during training. In this paper, we introduce a drastically different way to handle synthetic images that does not require seeing any real images at training time. Our approach builds on the observation that foreground and background classes are not affected in the same manner by the domain shift, and thus should be treated differently. In particular, the former should be handled in a detection-based manner to better account for the fact that, while their texture in synthetic images is not photo-realistic, their shape looks natural. Our experiments evidence the effectiveness of our approach on Cityscapes and CamVid with models trained on synthetic data only.
\end{abstract}

\section{Introduction}
\label{sec:intro}
As for many other computer vision tasks, deep networks have proven highly effective to perform semantic segmentation. Their main drawback, however, is their requirement for vast amounts of labeled data. In particular, acquiring such data for semantic segmentation is very expensive. For instance, pixel labeling of one Cityscapes image takes 90 minutes on average~\cite{CityScapes}. As a consequence, there has been a significant effort in the community to rely on the advances of computer graphics to generate synthetic datasets~\cite{GTAold,GTAnew,synthia}.

Despite the increasing realism of such synthetic data, there remain significant perceptual differences between synthetic and real images. Therefore, the performance of a state-of-the-art semantic segmentation network, such as~\cite{deeplab,FCN,PSP,deconv}, trained on synthetic data and tested on real images remains disappointingly low. While domain adaptation methods~\cite{ROAD,cycada,wild,curriculum,imageToimage,Crosscity} can improve such performance by explicitly accounting for the domain shift between real and synthetic data, they require having access to a large set of real images, albeit unsupervised, during training. As such, one cannot simply deploy a model trained off-line on synthetic data in a new, real-world environment.

\begin{figure}[t]
\centering
\scriptsize
\includegraphics[width=0.75\textwidth]{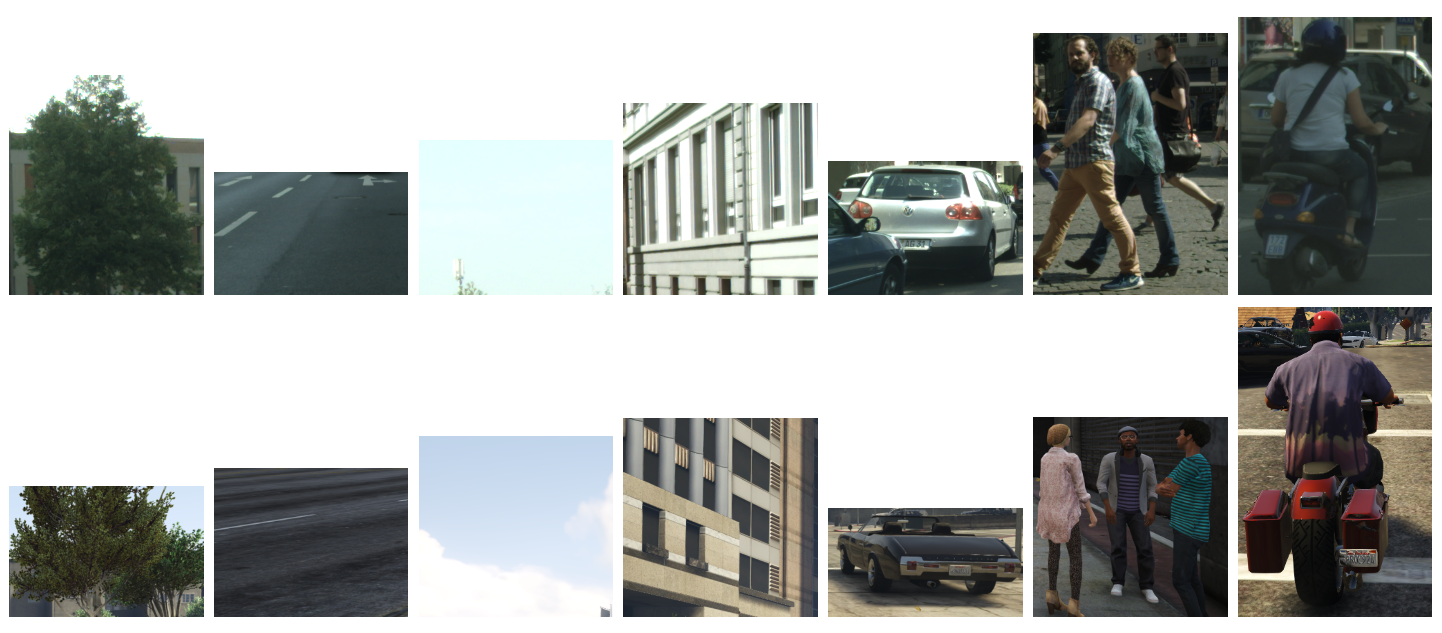}
{\scalebox{0.8}
{
\begin{tabular}{l | C{1.2cm} C{1.2cm} C{1.2cm}  C{1.2cm} C{1.2cm} C{1.2cm} C{1.2cm} C{1.2cm} C{1.2cm} C{1.2cm} } 
\hline
Methods & Traffic light & Traffic sign & Person& Rider & Car & Truck & Bus & Train & Motorcycle & Bicycle \\
 \hline
Segmentation &  22.3 & 23.8 & 48.7 & 13.3 & 75.1 & 14.3 & 21.2 & 2.1 & 24.2 & 7.3 \\
Detection-based &  26.7 & 42.5 & 52.2 & 28.5 & 76.2 & 19.6 & 31.6 &6.9 & 18.1 & 9.8 \\
\hline
\end{tabular}
}
}
\caption{Visual comparison of different classes in real Cityscapes images (Top) and synthetic GTA5 ones (Middle). Background classes (first 4 columns) are much less affected by the domain shift than foreground ones (last 3 columns), which present clearly noticeable differences in texture, but whose shape remain realistic. (Bottom)~We compare the accuracy of a semantic segmentation network (DeepLab) and of a detection-based model (Mask R-CNN), both trained on synthetic data only, on the foreground classes of Cityscapes. Note that the detection-based approach, by leveraging shape, yields significantly better results than the segmentation one.
} 
\label{fig:compare}

\end{figure}

In this paper, we introduce a drastically different approach to addressing the mismatch between real and synthetic data, based on the following observation: Not all classes suffer from the same type and degree of perceptual differences.
In particular, as can be seen in Fig.~\ref{fig:compare}, the \emph{texture} of background classes in synthetic images looks more realistic than that of foreground classes\footnote{We distinguish foreground classes from background ones primarily based on whether they have a well-defined shape and come in instances, or they are shapeless and identified by texture or material property. In essence, this corresponds to the distinction between {\it things} and {\it stuff} in~\cite{heitz}.
See Fig.~\ref{fig:compare} for examples.}. Nevertheless, the \emph{shape} of foreground objects in synthetic images looks very natural. We therefore argue that these two different kinds of classes should be treated differently. Specifically, we argue that semantic segmentation networks are well-suited to handle background classes because of their texture realism. By contrast, we expect object detectors to be more appropriate for foreground classes, particularly considering that modern detectors rely on generic object proposals. Indeed, when dealing with all possible texture variations of all foreground object classes, the main source of information to discriminate a foreground object from the background is shape. 

To empirically sustain our claim that detectors are better-suited for foreground classes, we trained separately a %state-of-the-art 
DeepLab~\cite{deeplab} semantic segmentation network and a Mask R-CNN~\cite{maskrcnn}, performing object detection followed by binary segmentation and class prediction, on synthetic data. At the bottom of Fig.~\ref{fig:compare}, we compare the mean Intersection over Union (mIoU) of these two models on the foreground classes of Cityscapes. Note that, except for {\it motorcycle}, the detector-based approach outperforms the semantic segmentation network on all classes.

Motivated by this observation, we therefore develop a simple, yet effective semantic segmentation framework that better leverages synthetic data during training. In essence, our model combines the foreground masks produced by Mask R-CNN with the pixel-wise predictions of the DeepLab semantic segmentation network. Our experiments on Cityscapes~\cite{CityScapes} and CamVid~\cite{camvid} demonstrate that this yields significantly higher segmentation accuracies on real data than only training a semantic segmentation network on synthetic data. Furthermore, our approach outperforms the state-of-the-art domain adaptation techniques~\cite{ROAD,cycada,wild,curriculum} 
without having seen any real images during training, and can be further improved by making use of unsupervised real images.

\begin{figure}[t]
\centering
\small
\begin{tabular}{c c }
\includegraphics[width=.45\textwidth]{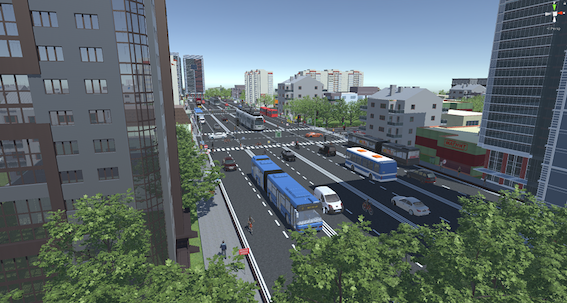} &
 \includegraphics[width=.45\textwidth]{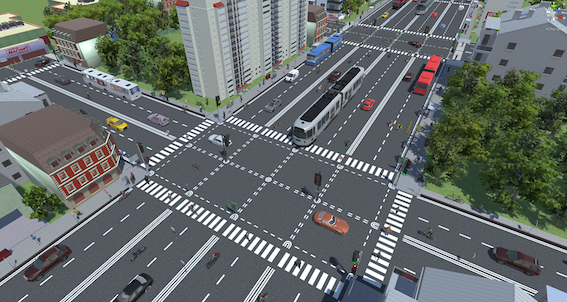}\\
\end{tabular}
\caption{Aerial views of our synthetic VEIS environment.} 
\label{fig:ourdataEnv}
\end{figure}

Furthermore, as a secondary contribution, we introduce a virtual environment created in the Unity3D framework, called \textbf{VEIS} (Virtual Environment for Instance Segmentation). This was motivated by the fact that existing synthetic datasets~\cite{synthia,GTAold,GTAnew} do not provide instance-level segmentation annotations for all the foreground classes of standard real datasets, such as CityScapes. VEIS automatically annotates synthetic images with instance-level segmentation for foreground classes. It captures urban scenes, such as those in Fig.~\ref{fig:ourdataEnv} shown from an aerial view, using a virtual camera mounted on a virtual car, yielding images such as those of Fig.~\ref{fig:ourdata}.
While not highly realistic, we show that, when used with a detector-based approach, this data allows us to boost semantic segmentation performance, despite it being of only little use in a standard semantic segmentation framework. We will make our data and the VEIS environment publicly available.

\section{Related work}
Semantic segmentation, that is, understanding an image at pixel-level, has been widely studied by the computer vision community~\cite{textonboost,superparsing,gouldmulti,mottaghi,farabet,recurrent,deephie,FCN,deconv,deeplab,conditional,PSP}. As for many other tasks, the most recent techniques rely on deep networks~\cite{deeplab,FCN,PSP,deconv}. Unfortunately, in contrast with image recognition problems, obtaining fully-supervised data for semantic segmentation, with pixel-level annotations, is very expensive and time-consuming. Two trends have therefore been investigated to overcome this limitation: Weakly-supervised methods and the use of synthetic data.

Weakly-supervised semantic segmentation aims to leverage a weaker form of annotation, such as image tags~\cite{SEC,weaklyandsemi,MIL,fromimgtopixel,saleh-eccv,salehincorporating,stc,wei2017object,oh2017exploiting}, bounding boxes~\cite{boxsup,boxeccv}, scribbles~\cite{whatsthepoint} and object size statistics~\cite{CCNN}, which are cheaper to obtain. While great progress has been made in this area, most existing methods focus only on foreground object classes and treat the background as one single entity. However, having detailed information about the different background classes is crucial in many practical scenarios, such as automated driving, where one needs to differentiate, e.g., the road from a grass field. To the best of our knowledge,~\cite{ICCV_saleh} constitutes the only method that considers multiple background classes for weakly-supervised semantic segmentation. This is achieved by leveraging both appearance and motion via a two-stream architecture trained using a loss based on classifier heatmaps. While this method is reasonably effective at segmenting background classes, there is still a huge gap compared to fully-supervised methods, especially in the foreground classes.

With the advance of computer graphics, generating fully-supervised synthetic data has become an attractive alternative to weakly-supervised learning. This has led to several datasets, such as SYNTHIA~\cite{synthia}, GTA5~\cite{GTAold} and VIPER~\cite{GTAnew}, as well as virtual environments to generate data~\cite{carla}. Unfortunately, despite the growing realism of such synthetic data, simply training a deep network on synthetic images to apply it to real ones still yields disappointing results. This problem is due to the domain shift between real and synthetic data, and has thus been tackled by domain adaptation methods~\cite{ROAD,cycada,wild,curriculum,imageToimage,Crosscity}, which, in essence, aim to reduce the gap between the feature distributions of the two domains. In~\cite{wild}, this is achieved by a domain adversarial training strategy inspired by the method of~\cite{ganin2015unsupervised,ganin2016domain}. This is further extended in~\cite{Crosscity} to align not only global, but also class-specific statistics. Domain adversarial training is combined in~\cite{ROAD} with a feature regularizer based on the notion of distillation~\cite{hinton2015distilling}. In~\cite{curriculum}, a curriculum style learning is introduced to align the label distribution over both entire images and superpixels. By contrast,~\cite{cycada} and~\cite{imageToimage} rely on a generative approach with cycle consistency to adapt the pixel-level and feature-level representations. 
While these methods outperform simply training a network on the synthetic data, without any form of adaptation, they all rely on having access to real images, without supervision, during training. As such, they cannot be directly deployed in a new environment without undergoing a new training phase.

Here, we follow an orthogonal approach to leverage synthetic data, based on the observation that foreground and background classes are subject to different perceptual mismatches between synthetic and real images. We therefore propose to rely on a standard semantic segmentation network for background classes, whose textures look quite realistic, and on a detection-based strategy for foreground objects because, while their textures look less natural, their shapes are realistic. Our experiments evidence that this outperforms state-of-the-art domain adaptation strategies. However, being orthogonal to domain adaptation, our method could also be used in conjunction with domain adaptation techniques. As a matter of fact,~\cite{BMVC}, which also argues that modern detectors rely on shape and discard the background texture, introduces a domain adaptation approach for the task of object detection, which could potentially be leveraged to deal with the foreground classes in our approach. We believe, however, that this goes beyond the scope of this paper. 

\section{Method}
\label{sec:method}
In this section, we introduce our approach to effectively use synthetic data for semantic segmentation in real driving scenarios. Note that, while we focus on driving scenarios, our approach generalizes to other semantic segmentation problems. However, synthetic data is typically easier to generate for urban scenes. Below, we first consider the case where we do not have access to any real images during training. We then introduce a simple strategy to leverage the availability of unsupervised real images.

\subsection{Detection-based Semantic Segmentation}
\label{sec:synth_only}

As discussed above, and illustrated by Fig.~\ref{fig:compare}, the perceptual differences of foreground and background classes in synthetic and real images are different. In fact, background classes in synthetic images look quite realistic, presenting very natural textures, whereas the texture of foreground classes does look synthetic, but their shape is realistic. We therefore propose to handle the background classes with a semantic segmentation network, but rather make use of a detection-based technique for the foreground classes. Below, we describe this in more detail, and then discuss how we perform semantic segmentation on a real image.\\

%\subsubsection{Dealing with Background Classes}

\begin{figure}[t]
\centering
\includegraphics[width=.8\textwidth]{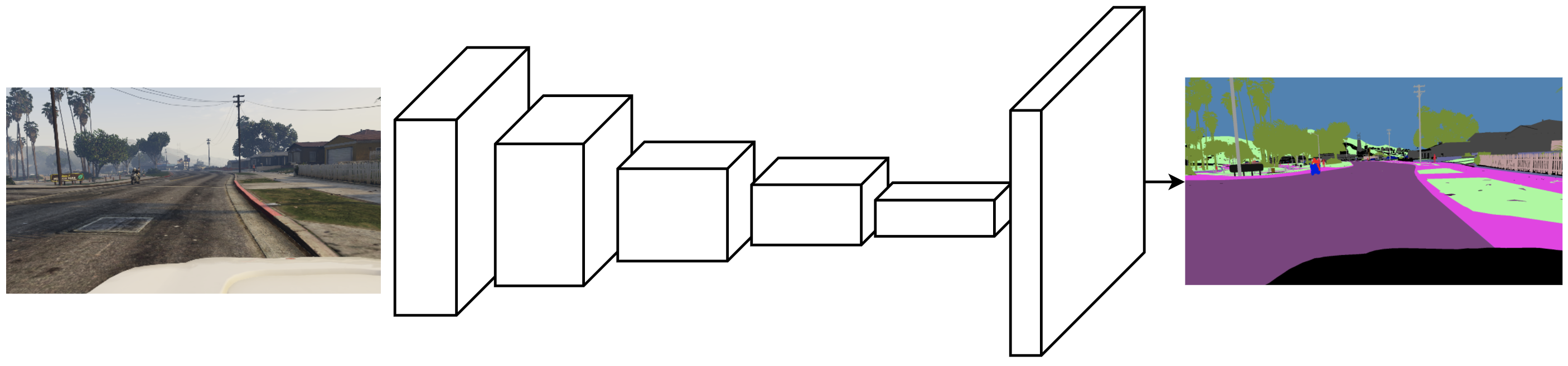}
\caption{{\bf Dealing with background classes.} We make use of the DeepLab semantic segmentation framework trained on synthetic GTA5~\cite{GTAold} frames with corresponding per-pixel annotations.}
\label{fig:GTA_Deeplab}
\end{figure}

\noindent\textbf{Dealing with Foreground Classes.}
To handle the background classes, we make use of the VGG16-based DeepLab model, depicted in Fig.~\ref{fig:GTA_Deeplab}. Specifically, we use DeepLab with a large field of view and dilated convolution layers~\cite{deeplab}. We train this model on the GTA5 dataset~\cite{GTAold} in which the background classes look photo-realistic. The choice of this dataset above others was also motivated by the fact that it contains all the classes of the commonly used real datasets, such as Cityscapes and CamVid. To train our model, we use the cross-entropy loss between the network's predictions and the ground-truth pixel-wise annotations of the synthetic images. Note that the network is trained on all classes, both foreground and background, but, as explained later, the foreground predictions are mostly discarded by our approach.

\begin{figure}[t]
\centering
\includegraphics[width=.8\textwidth]{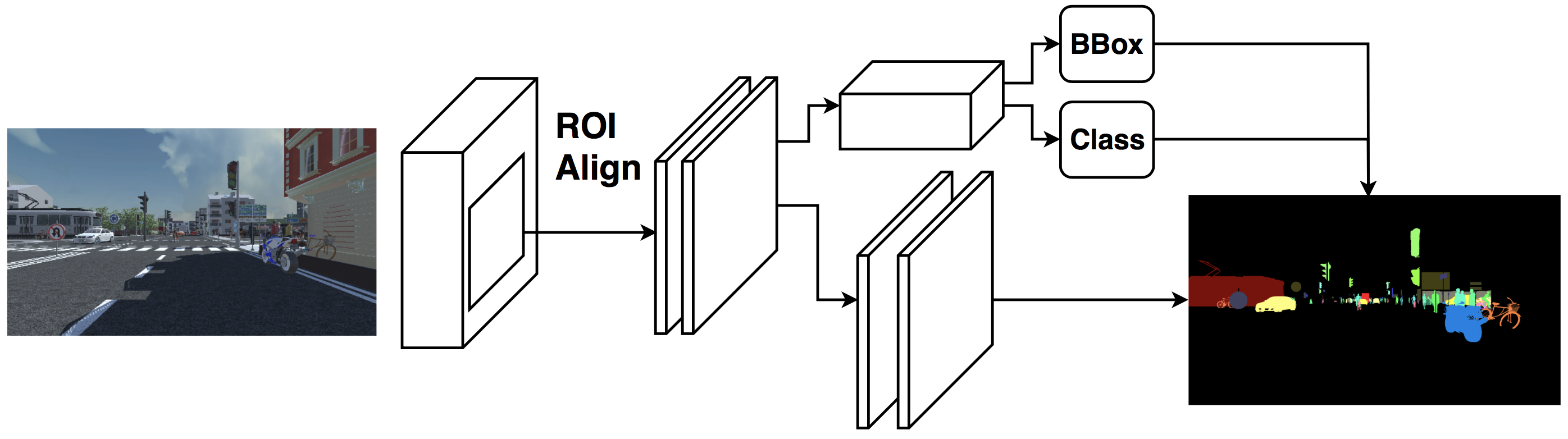}
\caption{{\bf Dealing with foreground classes.} We rely on the detection-based Mask R-CNN framework trained on our synthetic VEIS data with instance-level annotations. Note that these annotations were obtained automatically.}
\label{fig:maskrcnn}
\end{figure}
%\newline
%\newline
\noindent\textbf{Dealing with Foreground Classes.}
For foreground classes, our goal is to make use of a detection-based approach, which, as argued in Section~\ref{sec:intro}, relies more strongly on object shape than on texture, thus making texture realism of the synthetic data less crucial. Since our final goal is to produce a pixel-wise segmentation of the objects, we propose to rely on a detection-based instance-level semantic segmentation technique. Note that, once an object has been detected, segmenting it from the background within its bounding box is a comparatively easier task than semantic segmentation of an entire image. Therefore, texture realism is also not crucial here. To address this task, we make use of Mask R-CNN~\cite{maskrcnn}, which satisfies our criteria: As illustrated in Fig.~\ref{fig:maskrcnn}, it relies on an initial object detection stage, followed by a binary mask extraction together with object classification. Since existing synthetic datasets do not provide instance-level segmentations for all foreground classes of standard real datasets, we train Mask R-CNN using our own synthetic data, discussed in Section~\ref{sec:veis}. We make use of the standard architecture described in~\cite{maskrcnn}, as well as of the standard loss, which combines detection, segmentation, classification and regression terms. 

\begin{figure}[t]
\centering
\includegraphics[width=.8\textwidth]{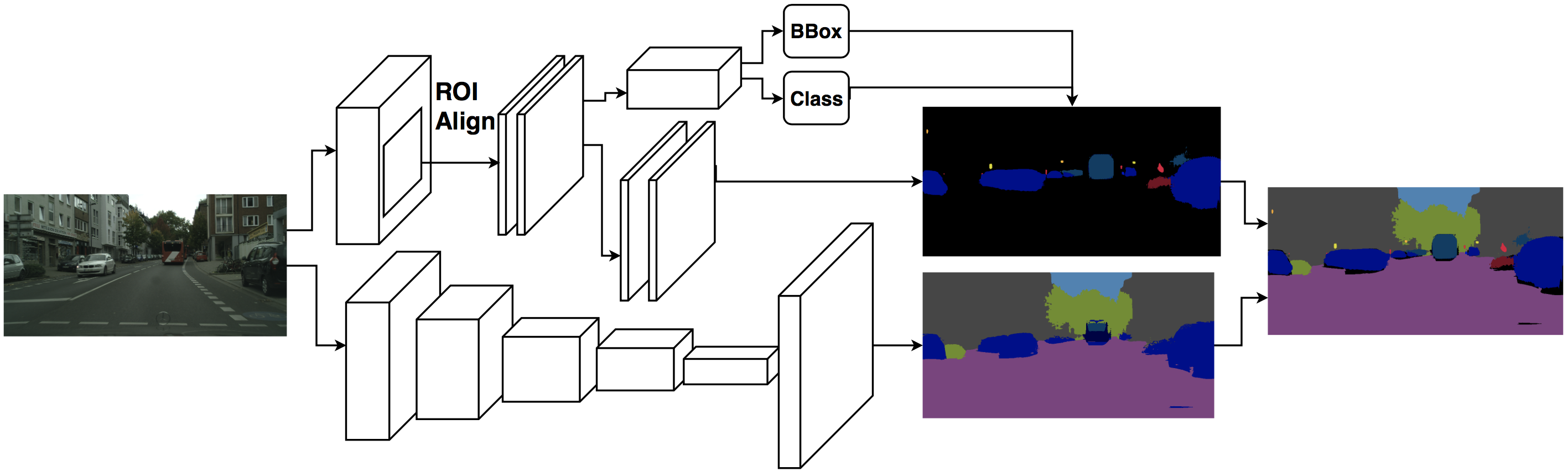}
\caption{{\bf Fusing foreground and background predictions.} Our approach combines the detection-based foreground predictions with the results of the semantic segmentation approach. Note that we do {\it not} require seeing any real images during training.}
\label{fig:method}
\end{figure}
%\newline
%\newline
\noindent\textbf{Prediction on Real Images.}
The two networks described above are trained using synthetic data only. At test time, we can then feed a real image to each network to obtain predictions. However, our goal is to obtain a single, pixel-wise semantic segmentation, not two separate kinds of outputs. To achieve this, as illustrated in Fig.~\ref{fig:method}, we fuse the two kinds of predictions, starting from the Mask R-CNN ones.
Specifically, given the Mask R-CNN predictions, we follow a strategy inspired by the panoptic segmentation procedure of~\cite{panoptic}, which constitutes an NMS-like approach to combine instance segments. More precisely, we first sort the predicted segments according to their confidence scores, and then iterate over this sorted list, starting from the most confident segment. If the current segment candidate overlaps with a previous segment, we remove the pixels in the overlapping region. The original procedure of~\cite{panoptic} relies on two different thresholds: One to discard the low-scoring segments and the other to discard non-overlapping yet too small segment regions. The values of these thresholds were obtained by grid search on real images. Since we do not have access to the ground-truth annotations of the real images, and in fact not even access to the real images during training, we ignore these two heuristics to discard segments, and thus consider all segments and all non-overlapping segment regions when combining the Mask R-CNN predictions.

Combining the Mask R-CNN predictions yield a semantic segmentation map that only contains foreground classes and has a large number of holes, where no foreground objects were found. To obtain our final semantic segmentation map, we fill these holes with the predictions obtained by the DeepLab network. That is, every pixel that is not already assigned to a foreground class takes the label with highest probability at that pixel location in the DeepLab result.

\subsection{Leveraging Unsupervised Real Images}
\label{sec:unsup}
The method described in Section~\ref{sec:synth_only} uses only synthetic images during training. In some scenarios, however, it is possible to have access to unlabeled real images at training time. This is in fact the assumption made by domain adaptation techniques. To extend our approach to this scenario, we propose to treat the predictions obtained by the method of Section~\ref{sec:synth_only} as pseudo ground-truth labels for the real images. To be precise, we make a small change to these predictions: In the holes left after combining the Mask R-CNN predictions, we assign the pixels that are predicted as foreground classes by the DeepLab model to an {\it ignore} label, so that they are not used for training. This is motivated by the fact that, as discussed above, the predictions of foreground classes by a standard semantic segmentation network are not reliable. We then use the resulting pseudo-labels as ground-truth to train a DeepLab semantic segmentation network from real images. As will be shown in our results, thanks to the good quality of our initial predictions, this helps further boost segmentation accuracy.

\section{The VEIS Environment and Dataset}
\label{sec:veis}

\begin{figure}[t]
\centering
\includegraphics[width=0.75\textwidth]{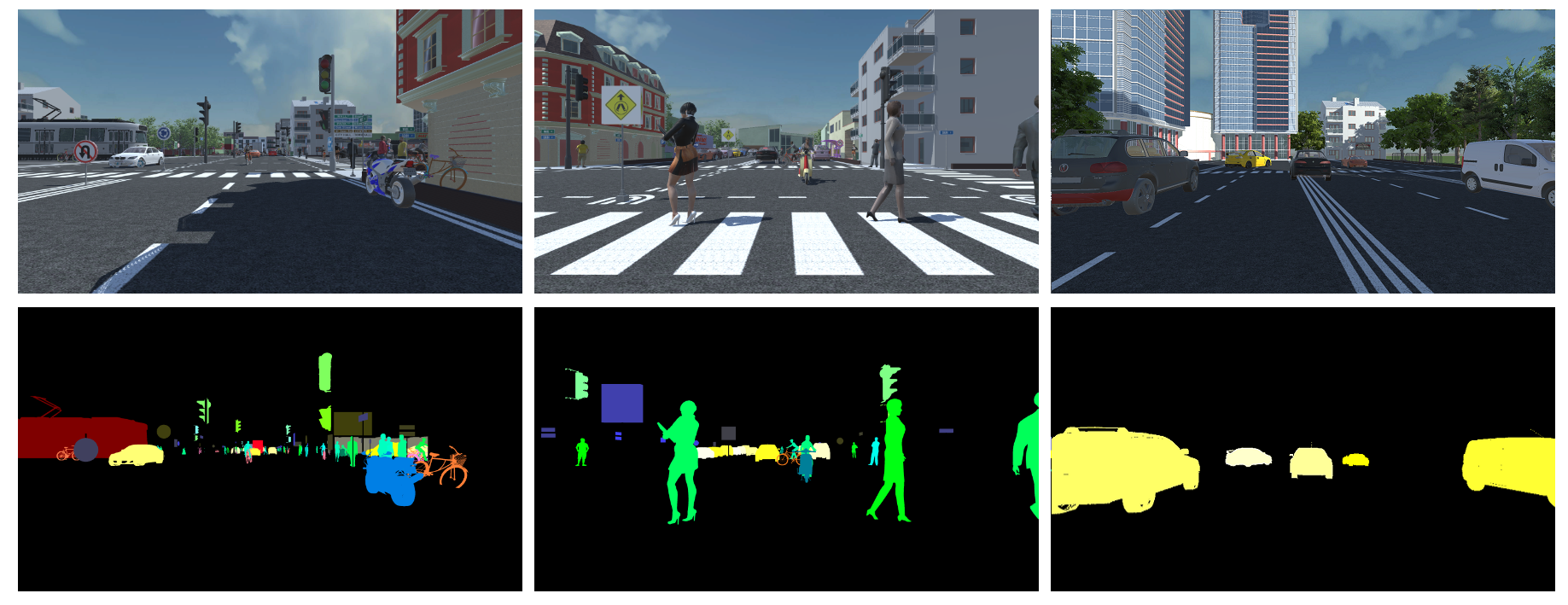}
\caption{
Example images and corresponding instance-level annotations, obtained automatically, from our synthetic VEIS dataset.} 
\label{fig:ourdata}
\end{figure}

In this section, we introduce our Virtual Environment for Instance Segmentation (VEIS) and the resulting dataset used in our experiments. While there are already a number of synthetic datasets for the task of semantic segmentation in urban scenes~\cite{synthia,GTAold,GTAnew}, they each suffer from some drawbacks. In particular, GTA5~\cite{GTAold} does not have instance-level annotations, and is thus not suitable for our purpose. By contrast, SYNTHIA~\cite{synthia} and VIPER~\cite{GTAnew} do have instance-level annotations, but not for all foreground classes of commonly-used real datasets, such as Cityscapes. For instance, {\it train}, {\it truck}, {\it traffic light} and {\it traffic sign} are missing in SYNTHIA, and {\it rider}, {\it traffic sign}, {\it train} and {\it bicycle} in VIPER. Furthermore,~\cite{GTAold,GTAnew} were acquired using the commercial game engine Grand Theft Auto V (GTAV), which only provides limited freedom for customization and control over the scenes 
%and objects 
to be captured, thus making it difficult to obtain a large diversity and good balance of classes. Obtaining ground-truth instance-level annotations in the GTAV game also involves a rather complicated procedure~\cite{GTAnew}.\\
%\newline
%\newline

\noindent\textbf{Environment.}
To alleviate these difficulties, we used the Unity3D~\cite{unity} game engine, in which one can manually design scenes with common urban structures and add freely-available 3D objects, representing foreground classes, to the scene. Example 3D scenes are shown in Fig.~\ref{fig:ourdataEnv}. Having access to the source code and manually constructing the scenes both facilitate generating annotations such as instance-level pixel-wise labels automatically. Specifically, before starting to generate the frames, our framework counts the number of instances of each class, and then assigns a unique ID to each instance. These unique IDs then automatically create unique textures and shaders for their corresponding instances. When data generation starts, both the original textures and shaders and the automatically created ones are rendered, thus allowing us to capture the synthetic image and the instance-level semantic segmentation map at the same time and in real time. Creating VEIS took 1 day to 1 person. This is very little effort, considering that VEIS allows us to have access to a virtually unlimited number of annotated images with the object classes of standard real urban scene datasets, such as CamVid and CityScapes.

As can be seen from the samples shown in Fig.~\ref{fig:ourdata}, the images generated by VEIS look less photo-realistic than those of~\cite{GTAold,GTAnew}. Therefore, as evidenced by our experiments, using them to train a semantic segmentation network does not significantly help improve accuracy on real images compared to using existing synthetic datasets. However, using these images within our proposed detection-based framework allows us to significantly improve semantic segmentation quality. This is due to the fact that, while not realistic in texture, the foreground objects generated by VEIS are realistic in shape, and our environment allowed us to cover a wide range of shape and pose variations.

Note that, in principle, we could have used other open source frameworks to generate our data, such as CARLA~\cite{carla}, implemented as an open-source layer of the Unreal Engine 4 (UE4)~\cite{unreal}. However, CARLA is somewhat too advanced for the purpose of our investigation. It targets the complete autonomous driving pipeline, with three different approaches covering a standard modular pipeline, an end-to-end approach based on imitation learning, and an end-to-end approach based on reinforcement learning. Since our goal was only to generate synthetic images covering a large diversity of foreground objects, we found Unity3D to be sufficient and easier to deploy.
\newline
\newline
\noindent\textbf{The VEIS Dataset.}
Using our VEIS environment, we generated images from two different types of scenes: 1) A multi-class, complex scene, where a city-like environment was synthesized with various objects of different classes. 2) A single-class, simple scene, where one or multiple objects of a single class were placed in a single road with background items (e.g., road, sidewalk, building, tree, sky), and images from multiple views were captured. Our VEIS dataset then contains 30180 frames from the multi-class scene and 31125 frames from the single-class scene, amounting to a total of 61305 frames with corresponding instance-level semantic segmentation. Note that the instance-level annotations were obtained with no human intervention. Some statistics of this dataset are shown in Table~\ref{tab:dataset}. In particular, we used a small amount of unique 3D objects for most of the classes and just repeated them in the scenes but with varying pose and articulation where applicable.

\section{Experiments}
In this section, we first describe the datasets used in our experiments and provide details about our learning and inference procedures. We then present the results of our model and compare it to state-of-the-art weakly-supervised semantic segmentation and domain adaptation methods.

\subsection{Datasets}
To train our model and the baseline, we make use of the synthetic GTA5 dataset~\cite{GTAold} and of our new VEIS dataset introduced in Section~\ref{sec:veis}. Furthermore, we also provide results of fully-supervised models trained on the synthetic SYNTHIA~\cite{synthia} and VIPER~\cite{GTAnew} datasets. At test time, we evaluate the models on the real images of the CityScapes~\cite{CityScapes} and CamVid~\cite{camvid} road scene datasets.
Below, we briefly discuss the characteristics of these datasets.

\begin{table}[t]
\renewcommand{\arraystretch}{1.1}
\small
\centering
\caption{Some statistics of our synthetic data}
\scriptsize
\label{tab:dataset}
 \setlength\tabcolsep{1.1pt} % default value: 6pt
 \scalebox{.9}
 {
\begin{tabular}{l|c@{ }@{ }  @{ } c@{ }@{ }  @{ } c@{ }@{ }  @{ } c@{ }@{ }  @{ } c@{ }@{ }  @{ } c@{ }@{ }  @{ } c@{ }@{ }  @{ } c@{ }@{ }  @{ } c@{ }@{ }  @{ } c }
 Class  & \rotatebox[origin=l]{70}{T. light} & \rotatebox[origin=l]{70}{T. sign}  & \rotatebox[origin=l]{70}{Person} & \rotatebox[origin=l]{70}{Rider} & \rotatebox[origin=l]{70}{Car} & \rotatebox[origin=l]{70}{Truck} & \rotatebox[origin=l]{70}{Bus} & \rotatebox[origin=l]{70}{Train} & \rotatebox[origin=l]{70}{M. bike} & \rotatebox[origin=l]{70}{Bicycle} \\
 \hline
 \#unique instances & 3 & 69 & 31 & 1 & 13 & 6 & 3 & 3 & 7 & 4\\
 \#instances in dataset & 101771& 261015 & 176552 & 67073 & 148760 & 26847 & 45082 & 12071 & 50687 & 67672\\
\end{tabular}
}
\end{table}

\noindent\textbf{GTA5~\cite{GTAold}} was captured using the Grand Theft Auto V video game and contains 24966 photo-realistic images with corresponding pixel-level annotations. The resolution of the images is $1920\times 1080$ and the class definitions of the semantic categories are compatible with those in the Cityscapes dataset.

\noindent\textbf{VIPER~\cite{GTAnew}} is a slightly more recent dataset than GTA5, also acquired using the Grand Theft Auto V video game, but covering a wider range of weather conditions. It contains more than 250K high-resolution (1920$\times$1080) video frames, all annotated with ground-truth labels for both
low-level and high-level vision tasks, including optical flow, semantic instance segmentation, object detection and tracking, object-level 3D scene layout, and visual odometry. In our experiments, the model exploiting VIPER was trained using the training and validation sets of this dataset (over 180K frames). While VIPER is larger than GTA5, its labels are not really compatible with Cityscapes. For example, the classes {\it rider} and {\it wall} are missing; the class {\it pole} has been incorporated into {\it infrastructure}\footnote{To evaluate the \emph{pole} class, we considered any \emph{infrastructure} prediction as \emph{pole}, which is the dominant label in this slightly broader class.}; the windows of the cars are not labeled as {\it car} unlike in Cityscapes. This explains why most of our experiments rather rely on GTA5.

\noindent\textbf{SYNTHIA~\cite{synthia}}
is another dataset of synthetic images, with a subset called SYNTHIA-RAND-CITYSCAPES meant to be compatible with Cityscapes. This subset contains 9,400 images with pixel-level semantic annotations. However, some classes, such as {\it train}, {\it truck} and {\it terrain}, have no annotations. 
As for VIPER, we show the performance of a fully-supervised method trained on SYNTHIA. This is for the sake of completeness, even though we favor GTA5 since it contains all the classes of Cityscapes.

\noindent\textbf{Cityscapes~\cite{CityScapes}} is a large-scale dataset of real images, containing high-quality pixel-level annotations for 5000 images collected in street scenes from 50 different cities. There is also another set of images with coarse level annotations. 
We report the results of all models on the 500 validation images. Furthermore, the methods that rely on unsupervised real images during training, including ours,
%approach of Section~\ref{sec:unsup}, 
were trained using the 22971 train/train-extra  RGB frames of this dataset. 

\noindent\textbf{CamVid~\cite{camvid}} 
consists of over 10 minutes of high quality 30 Hz footage. The videos were captured at $960 \times 720$ resolution with a camera mounted inside a car. Three of the four sequences were shot in daylight, and the fourth one was captured at dusk. This dataset contains 32 categories. In our experiments, following~\cite{camvid}, we used a subset of 11 classes. The dataset is split into 367 training, 101 validation and 233 test images. Note that, as for the Cityscapes dataset, we evaluate on the test set and, when training on unsupervised data, used the RGB frames of training+validation without any type of annotation. 

\subsection{Implementation Details}

As discussed in Section~\ref{sec:method}, our approach makes use of two types of networks: DeepLab (Large FOV)~\cite{deeplab} for semantic segmentation and Mask R-CNN~\cite{maskrcnn} for instance-level segmentation. Below, we briefly discuss these models.
\newline
\newline
\noindent\textbf{DeepLab.} 
To train our semantic segmentation networks, using either the synthetic GTA5 dataset or real images with pseudo ground truth, we used a DeepLab model with a large field of view and dilated convolution layers. We relied on stochastic gradient descent with a learning rate starting at $25\times 10^{-5}$, with a decrease factor of $10$ every $40k$ iterations, a momentum of $0.9$, a weight decay of $0.0005$, and mini-batches of size $1$. Similarly to recent %semantic segmentation 
methods~\cite{ICCV_saleh,deeplab,FCN,PSP}, the weights of our semantic segmentation network were initialized with those of the VGG-16 classifier~\cite{vgg} pre-trained on ImageNet~\cite{ILSVRC}. Note that, because of limited GPU memory, we down-sampled the high resolution images of Cityscapes, GTA5, VIPER, and SYNTHIA by a factor 2 when using them for training.
\newline
\newline
\noindent\textbf{Mask R-CNN.}
To train a Mask R-CNN network, we make use of the implementation provided by the ``Detectron" framework~\cite{detectron}. We train an end-to-end Mask R-CNN model with a 64 $\times$ 4d ResNeXt-101-FPN backbone, pre-trained on ImageNet, on our synthetic VEIS dataset. We use mini-batches of size 1 and train the model for 200k iterations, starting with a learning rate of 0.001 and reducing it to 0.0001 after 100k iterations. 

\subsection{Evaluated Methods}

In our experiments, we report the results of the following methods:
\newline\noindent \textbf{GTA5~\cite{ROAD}:} This baseline denotes a DeepLab model trained on GTA5 by the authors of~\cite{ROAD}. We directly report the numbers as provided in~\cite{ROAD}.
\newline\noindent \textbf{GTA5:} This corresponds to our replication of the baseline above. We found our implementation to yield an average accuracy 9.4\% higher than the one reported in~\cite{ROAD}. As such, this constitutes our true baseline.
\newline\noindent \textbf{SYNTHIA:} This refers to a DeepLab model trained on the SYNTHIA~\cite{synthia} dataset instead of GTA5. 
\newline\noindent \textbf{VIPER:} This baseline denotes a DeepLab model trained on the VIPER dataset. 
\newline\noindent \textbf{VEIS:} This corresponds to training a DeepLab model on our new dataset. Note that here we considered all the classes, both foreground and background ones, for semantic segmentation, ignoring the notion of instances.
\newline\noindent \textbf{GTA5+VEIS:} This denotes a DeepLab model trained jointly on GTA5 and our new dataset for semantic segmentation. 
\newline\noindent \textbf{GTA5+VEIS and Pseudo-GT:} For this baseline, we used the results of the GTA5+VEIS baseline to generate pseudo-labels on the real images. We then trained another DeepLab network using these pseudo-labels as ground-truth. In essence, this corresponds to the approach discussed in Section~\ref{sec:unsup}, but without handling the foreground classes in a detection-based manner. 
\newline\noindent \textbf{Ours:} This corresponds to our method in Section~\ref{sec:synth_only}, which relies on the GTA5 synthetic data and uses a detection-based model for foreground classes combined with a DeepLab semantic segmentation network for the background ones.
\newline\noindent \textbf{Ours and Pseudo-GT:} This consists of using the method above (Ours) to generate pseudo-labels on the real images, and training a DeepLab model from these pseudo-labels, as introduced in Section~\ref{sec:unsup}.

\subsection{Experimental Results}
We now compare the results of the different methods discussed above on the real images of Cityscapes and CamVid. Furthermore, we also compare our approach to the state-of-the-art weakly supervised semantic segmentation and domain adaptation methods on Cityscapes.

\begin{figure}[!t]
\centering
\tiny
\begin{tabular}{c c c c c}
Image & Ground truth & GTA5+VEIS & Ours & Ours ps-GT \\
\includegraphics[width=.182\textwidth]{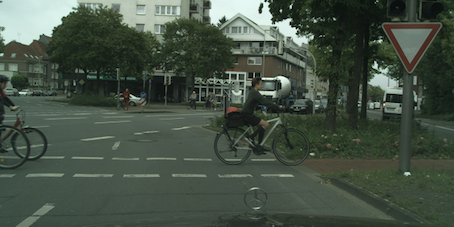}&
\includegraphics[width=.182\textwidth]{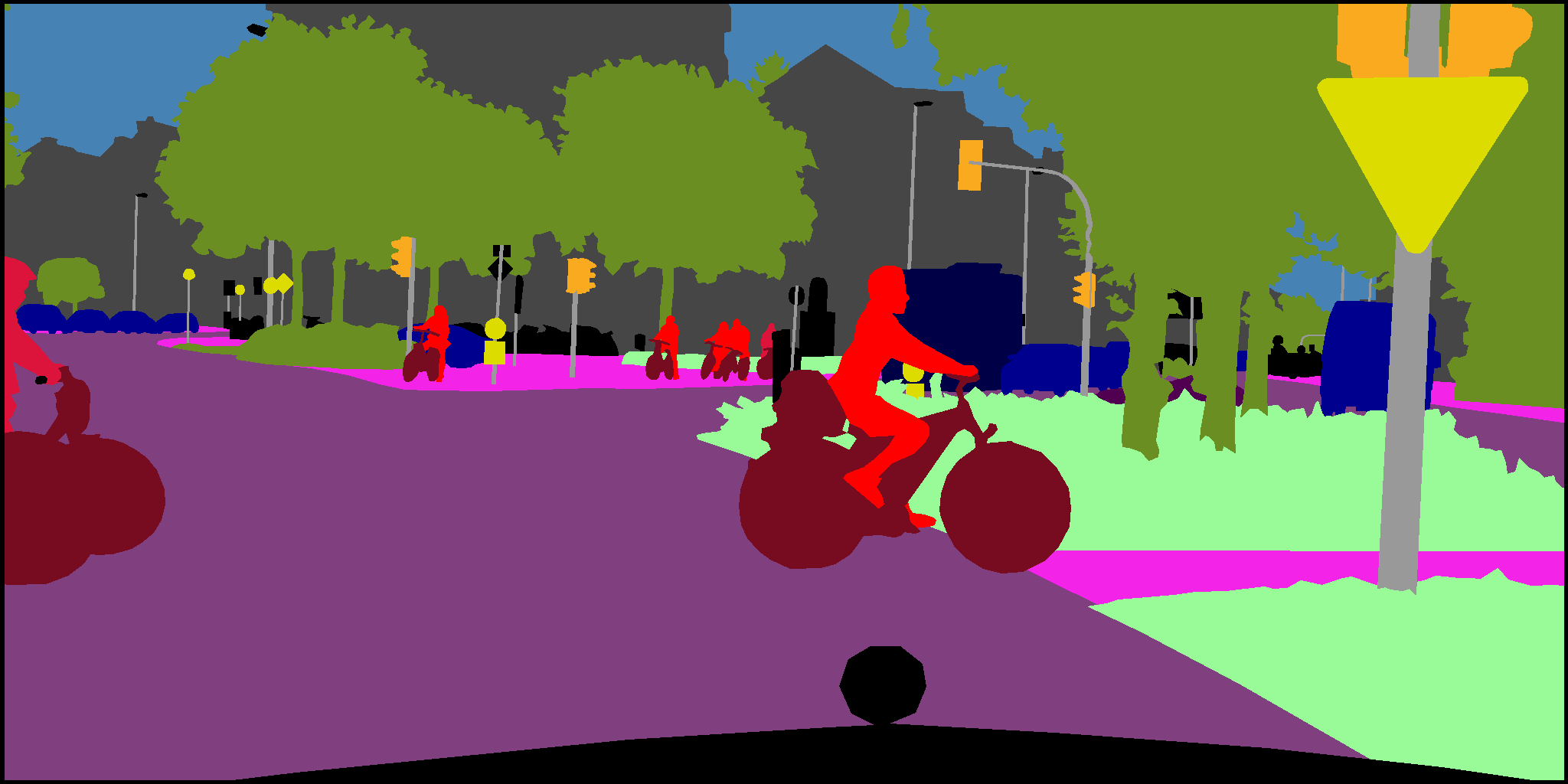} &
\includegraphics[width=.182\textwidth]{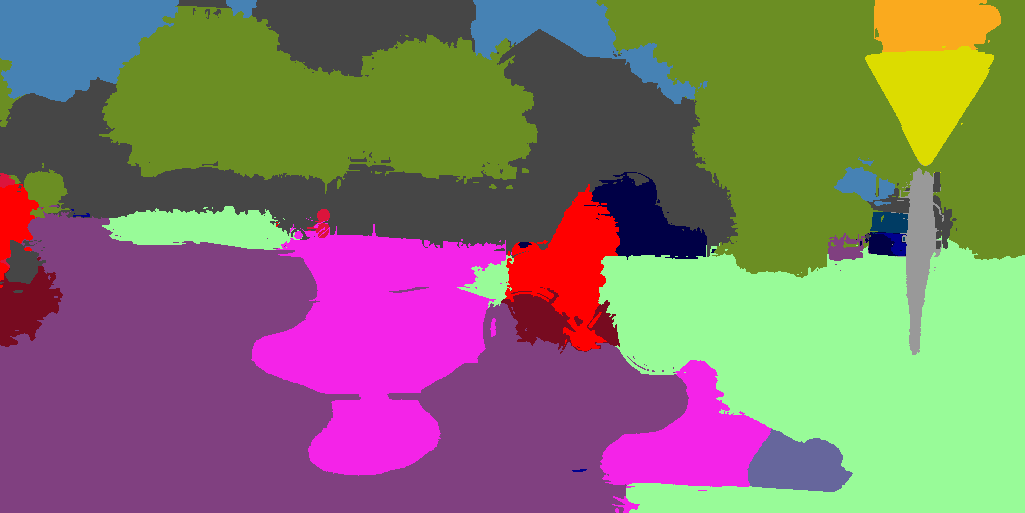} &
\includegraphics[width=.182\textwidth]{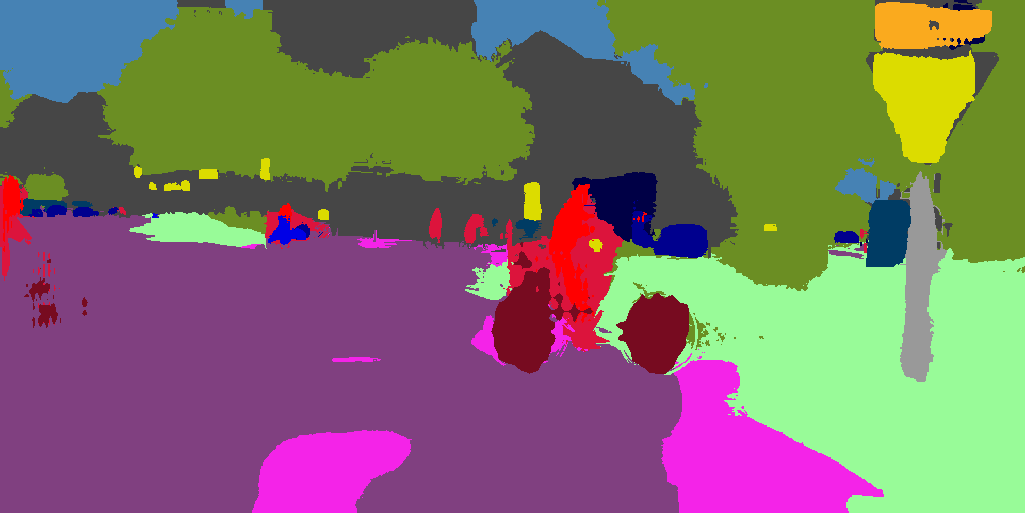} &
\includegraphics[width=.182\textwidth]{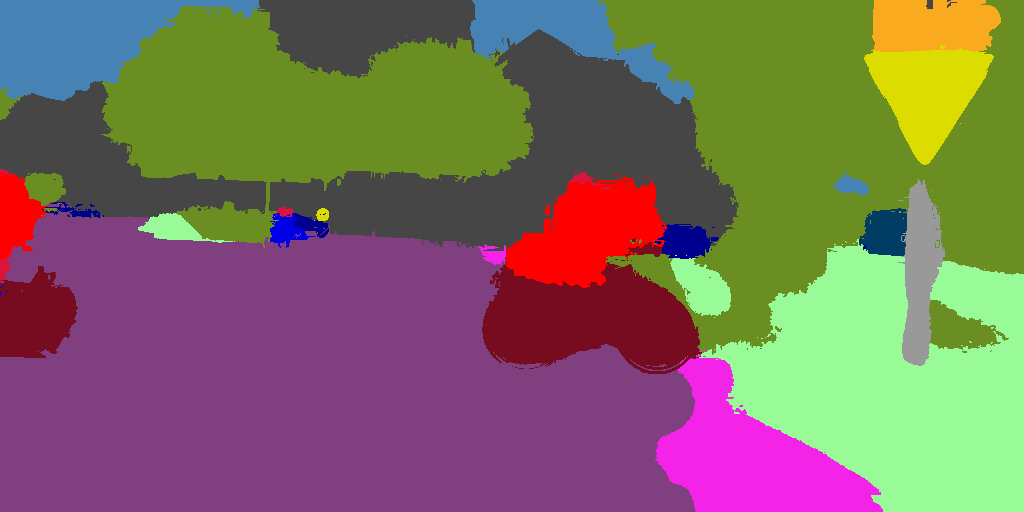} \\
\includegraphics[width=.182\textwidth]{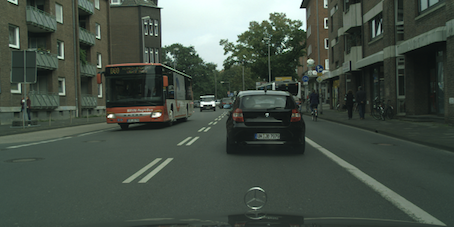}&
\includegraphics[width=.182\textwidth]{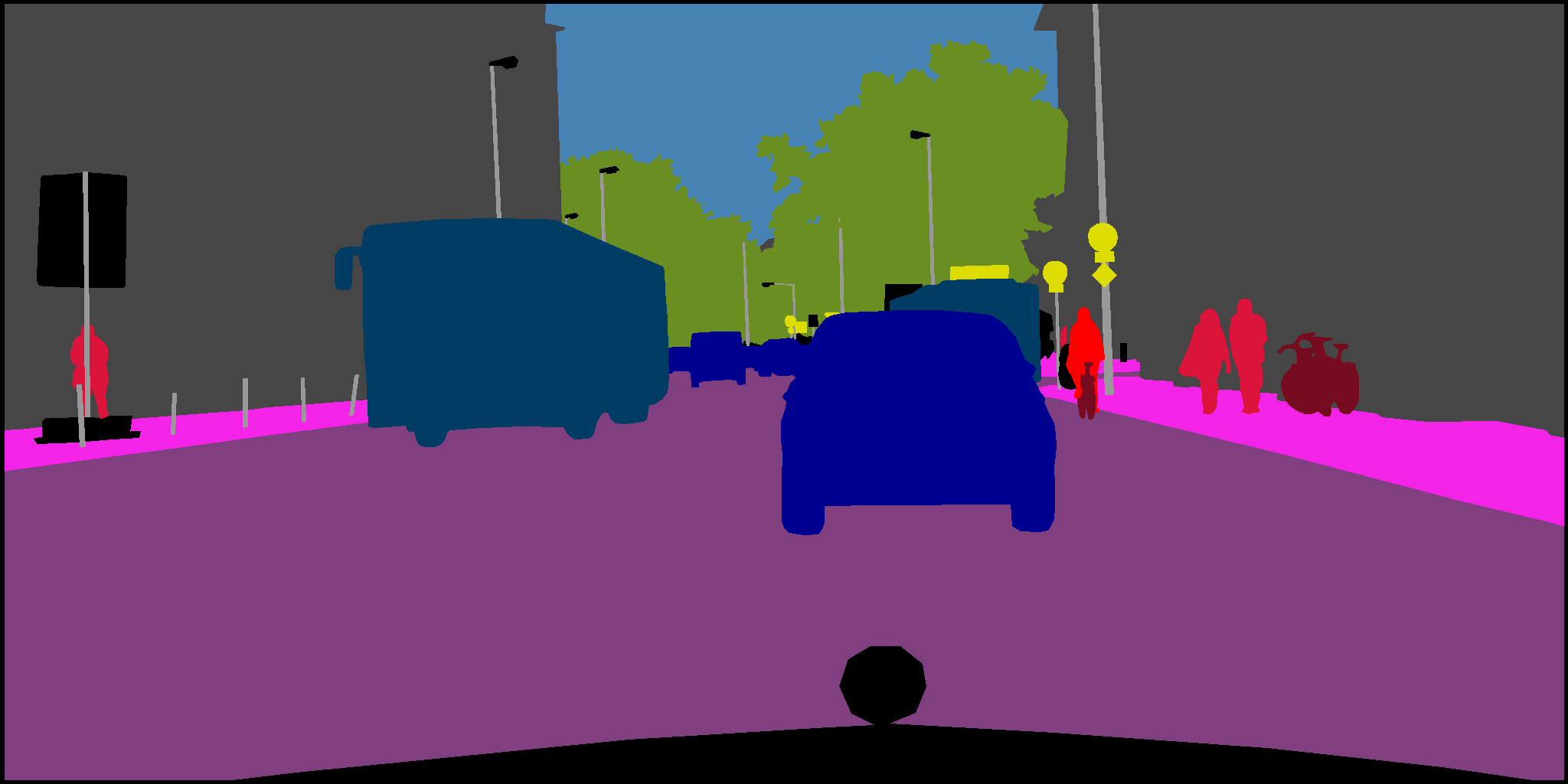} &
\includegraphics[width=.182\textwidth]{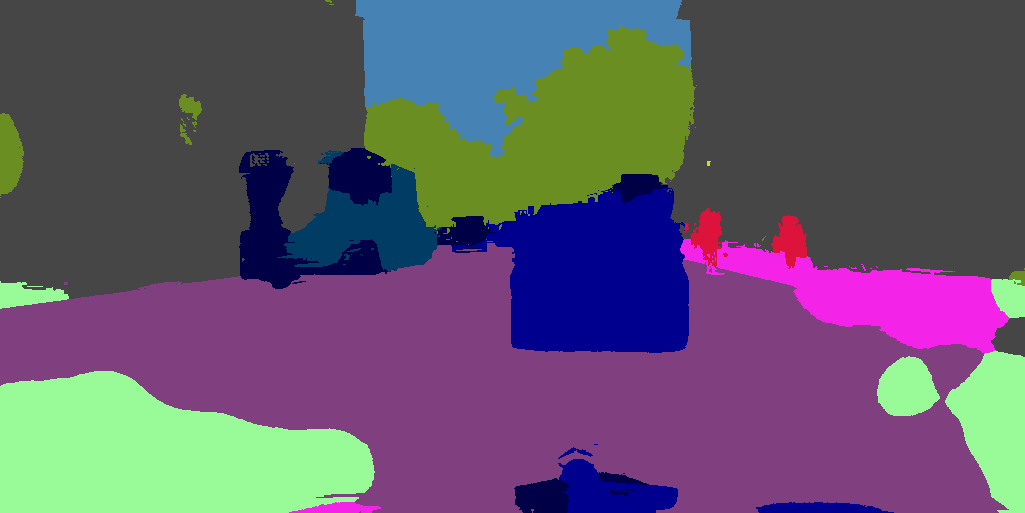} &
\includegraphics[width=.182\textwidth]{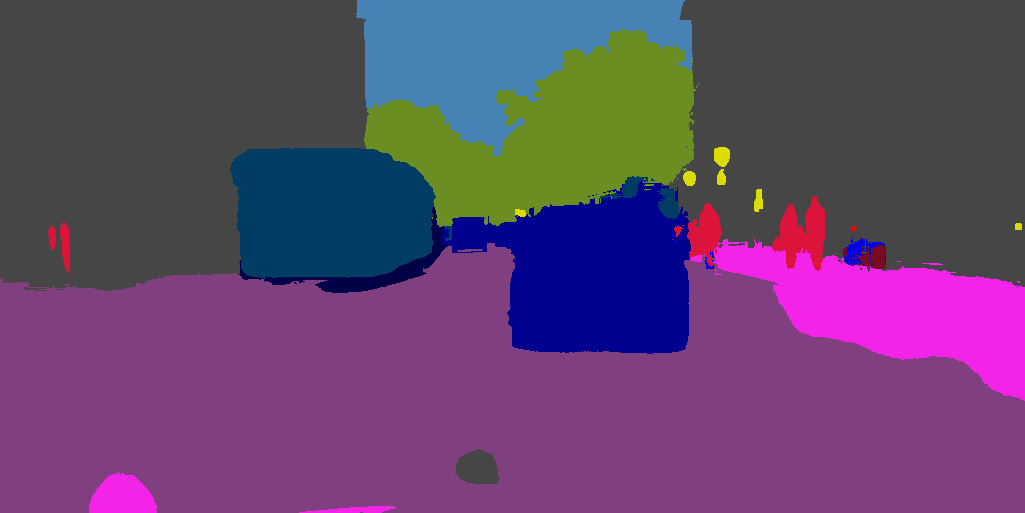} &
\includegraphics[width=.182\textwidth]{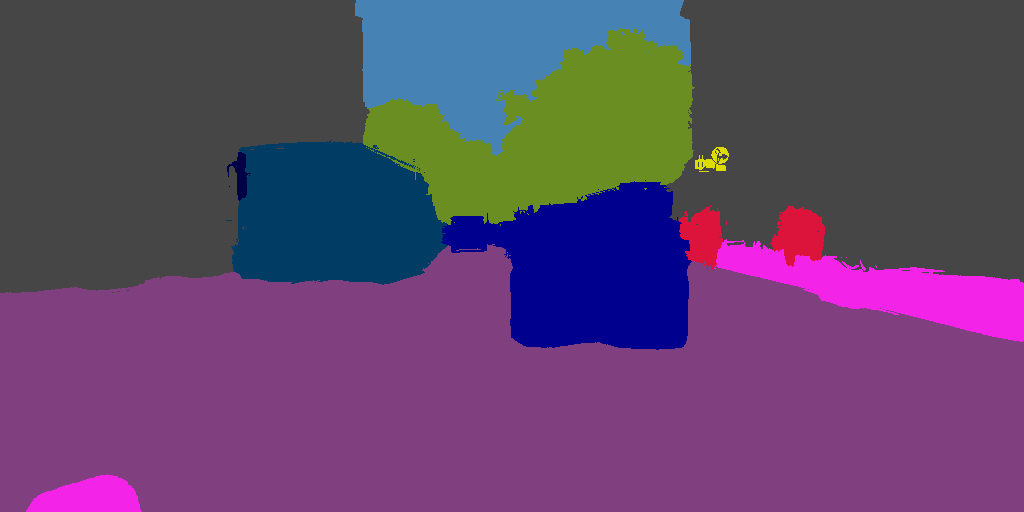} \\
\includegraphics[width=.182\textwidth]{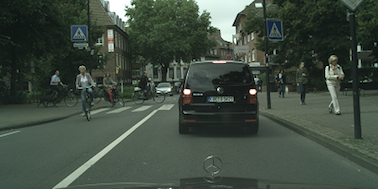}&
\includegraphics[width=.182\textwidth]{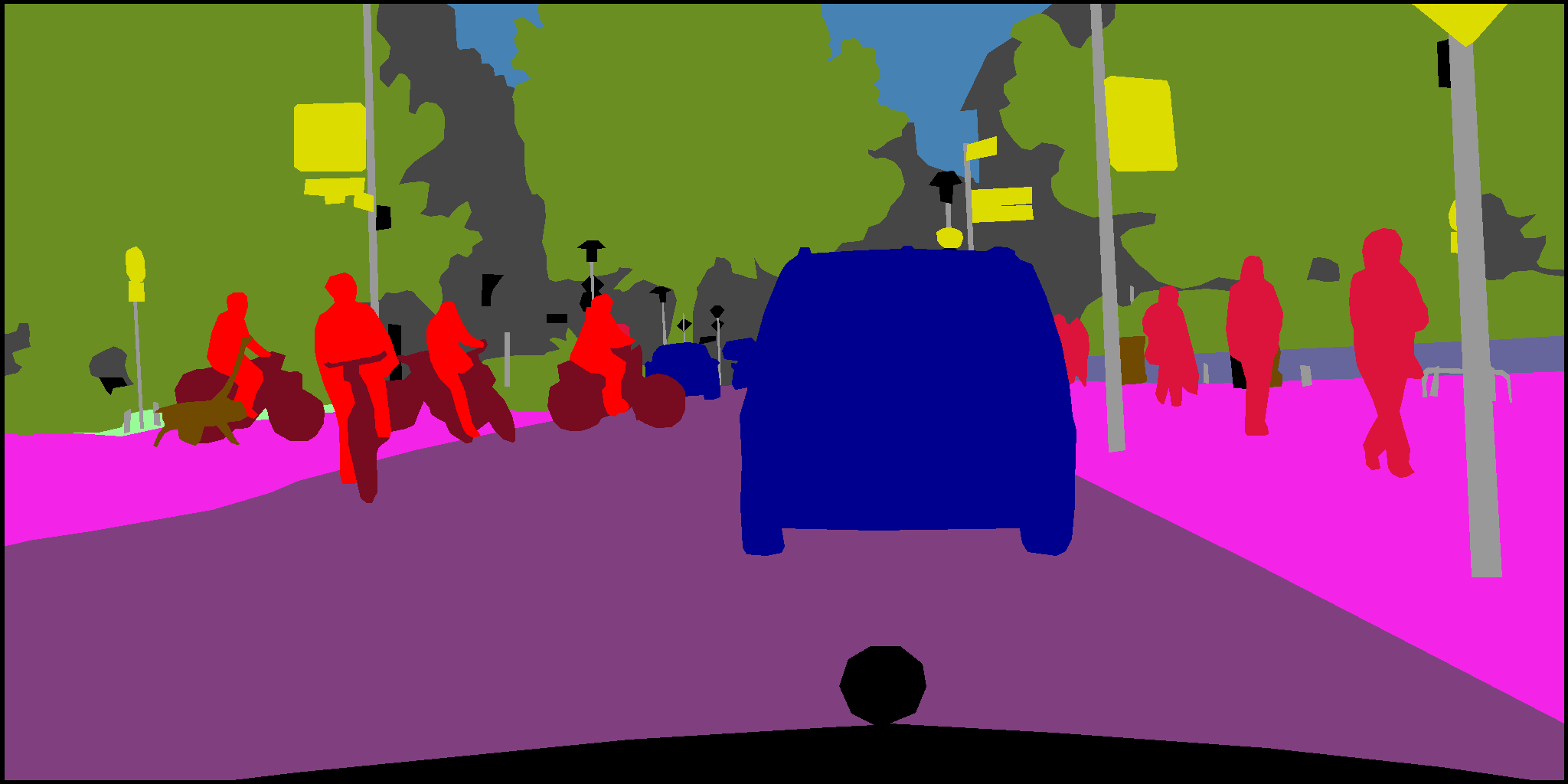} &
\includegraphics[width=.182\textwidth]{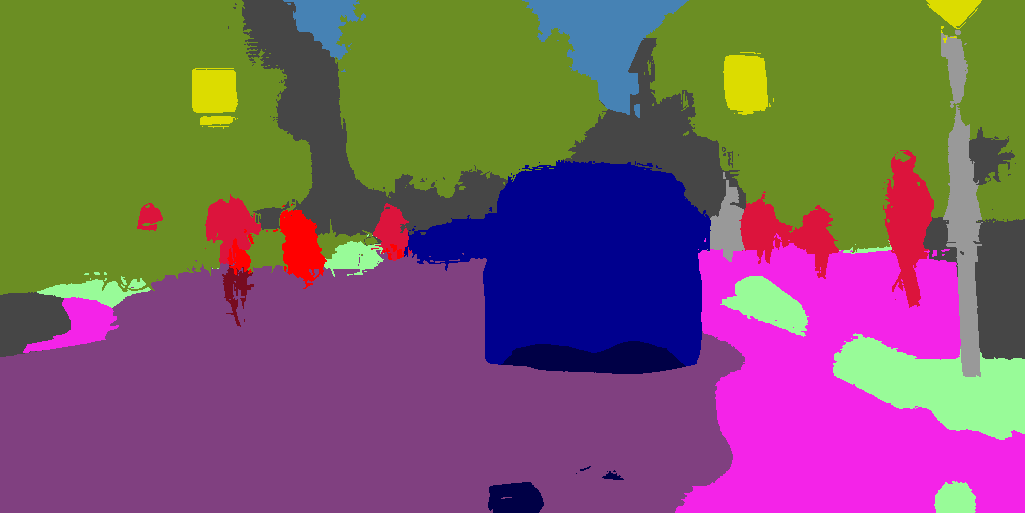} &
\includegraphics[width=.182\textwidth]{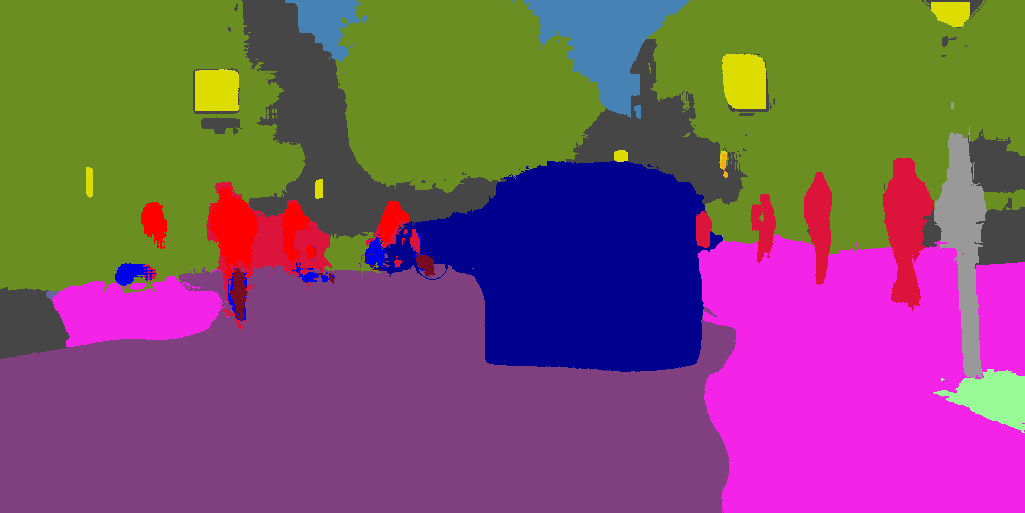} &
\includegraphics[width=.182\textwidth]{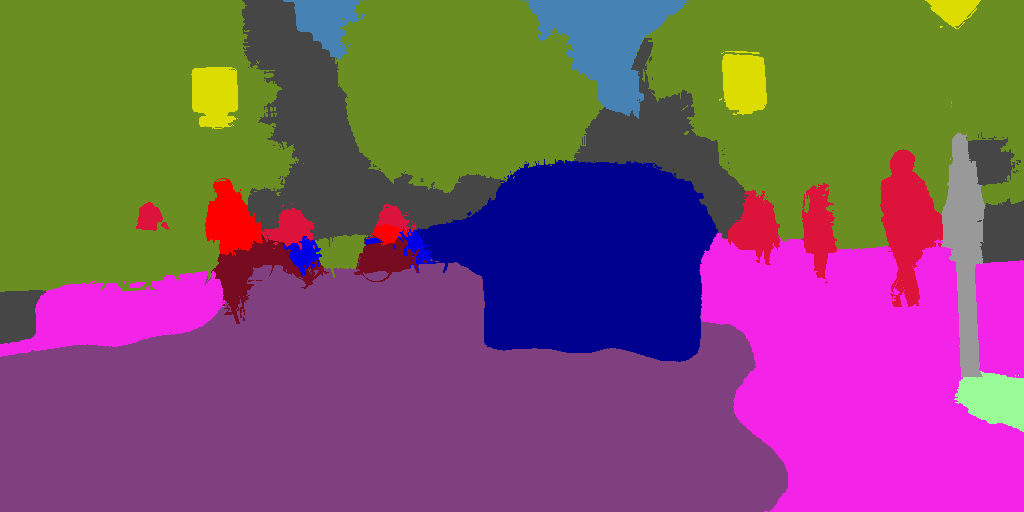} \\

\end{tabular}
\caption{Qualitative results on Cityscapes.} 
\label{fig:ourresults}
\end{figure}

In Table~\ref{tab:setups}, we provide the results of the methods described above on Cityscapes. The foreground classes are highlighted. In essence, we can see that GTA5 performs better than training DeepLab on the datasets \{SYNTHIA,VIPER,VEIS\} alone, because these datasets either do not contain all the Cityscapes classes \{SYNTHIA,VIPER\}, or because they are less realistic \{VEIS\}. Complementing GTA5 with VEIS \{GTA5+VEIS\} improves the results by only a small margin, again because of the non-photo-realistic VEIS images. By contrast, using GTA5 and VEIS jointly within our approach (Ours) yields a significant improvement. This is because our detection-based way of dealing with foreground classes is less sensitive to photo-realism, but focuses on shape, which does look natural in our VEIS data. As a matter of fact, our improvement is particularly marked for foreground classes. Finally, while using pseudo-labels from the \{GTA5+VEIS\} baseline only yields a minor improvement, their use within our framework gives a significant accuracy boost. Some qualitative results are shown in Fig.~\ref{fig:ourresults}.

In Table~\ref{tab:mask_eval}, we compare our approach with the state-of-the-are weakly-supervised method of~\cite{ICCV_saleh} and with state-of-the-art domain adaptation methods. The results for these methods were directly taken from their respective papers.
Note that, even without seeing the Cityscapes images at all, our approach (Ours) outperforms all these baselines. Using unsupervised Cityscapes images (Ours+pseudo-GT) helps to further improve over the baselines. To also show that our approach generalizes to the situation where there is no domain shift, we trained its two components on Cityscapes and evaluated it on the Cityscapes validation set (Fully Sup. Ours in Table~\ref{tab:mask_eval}). This significantly improves the segmentation accuracy of foreground classes (e.g., person, car, truck, bus, train, motorbike, bike).

\begin{table}[t]
\renewcommand{\arraystretch}{1.4}
\small
\centering
\caption{{\bf Comparison of models trained on synthetic data.} All the results are reported on the Cityscapes validation set. Note that ps-GT (pseudo-GT) indicates the use of unlabeled real images during training.}
\label{tab:setups}
\tiny
% \qquad  % get some separation between the two tabulars
 \setlength\tabcolsep{1.5pt} % default value: 6pt
 \scalebox{.93}
 {
\begin{tabular}{l|c c c c c c a a c c c a a a a a a a a |c }
  & \rotatebox[origin=l]{90}{road} &\rotatebox[origin=l]{90}{side.} & \rotatebox[origin=l]{90}{buil.} & \rotatebox[origin=l]{90}{wall} & \rotatebox[origin=l]{90}{fence} & \rotatebox[origin=l]{90}{pole} & \rotatebox[origin=l]{90}{light} & \rotatebox[origin=l]{90}{sign} & \rotatebox[origin=l]{90}{Vege.} & \rotatebox[origin=l]{90}{terr.} & \rotatebox[origin=l]{90}{sky} & \rotatebox[origin=l]{90}{person} & \rotatebox[origin=l]{90}{rider} & \rotatebox[origin=l]{90}{car} & \rotatebox[origin=l]{90}{truck} & \rotatebox[origin=l]{90}{bus} & \rotatebox[origin=l]{90}{train} & \rotatebox[origin=l]{90}{motor} & \rotatebox[origin=l]{90}{bike} & \rotatebox[origin=l]{90}{mIOU} \\
\hline
GTA5~\cite{ROAD}& 29.8 & 16.0 & 56.6 &  9.2 & 17.3 & 13.5 & 13.6 & 9.8 & 74.9 & 6.7 & 54.3 & 41.9 &  2.9 & 45.0 & 3.3 & 13.1 & 1.3 & 6.0 & 0.0 & 21.9\\
GTA5 & 80.5 & 26.0 & 74.7 & 23.0 & 9.8 & 9.1 & 13.4 & 7.3 & 79.4 & 28.6 & 72.1 & 40.4 & 5.1 & 77.8 & 23.0 & 18.6 & 1.2 & 5.3 & 0.0 & 31.3\\
SYNTHIA& 36.7 & 22.7 & 51.0 & 0.3 & 0.1 & 16.6 & 0.1 & 9.5 & 72.5 & 0.0 & 78.4 & 47.5 & 5.6 & 61.4 & 0.0 & 13.0 & 0.0 & 3.2 & 3.1 & 22.1\\
VIPER& 36.9 & 19.0 & 74.7 & 0.0 & 5.3 & 7.1 & 10.0 & 10.1 & 78.7 & 13.6 & 69.6 & 43.0 & 0.0 & 41.2 & 20.8 & 13.9 & 0.0 & 9.1 & 0.0 & 23.9\\
VEIS& 70.8 & 9.5 & 50.9 & 0.0 & 0.0 & 0.3 & 15.6 & 26.8 & 66.8 & 12.7 & 52.3 & 44.0 & 14.2 & 60.6 & 10.2 & 8.2 & 3.2 & 5.5 & 11.8 & 24.4\\
GTA5+VEIS& 66.2 & 21.6 & 72.3 & 15.7 & 18.3 & 12.3 & 22.3 & 23.8 & 78.4 & 11.3 & 74.6 & 48.7 & 13.3 & 75.1 & 14.3 & 21.2 & 2.1 & 24.2 & 7.3 & 32.8\\
GTA5+VEIS\&ps-GT& 77.6 & 26.8 & 75.5 & 19.4 & 19.5 & 4.8 & 18.7 & 19.8 & 79.5 & 21.7 & 78.9 & 47.3 & 8.7 & 77.6 & 23.1 & 16.1 & 2.2 & 15.6 & 0.0 & 33.3\\
Ours&71.9 & 23.8 & 75.5 & 23.4 & 14.9 & 9.3 & 26.7 & 42.5 & 80.1 & 34.0 & 76.3 & 52.2 & 28.5 & 76.2 & 19.6 & 31.6 & 6.9 & 18.1 & 9.8 & 38.0\\
Ours\&ps-GT&79.8 & 29.3 & 77.8 & 24.2 & 21.6 & 6.9 & 23.5 & 44.2 & 80.5 & 38.0 & 76.2 & 52.7 & 22.2 & 83.0 & 32.3 & 41.3 & 27.0 & 19.3 & 27.7 & 42.5\\
\end{tabular}
}

\end{table}

The results on CamVid in Table~\ref{tab:camvid}, where we compare our method to fully-supervised techniques that make use of CamVid images and annotations to train a model, GTA5-based baselines, and the state-of-the-art weakly-supervised method, show a similar trend. Our approach clearly outperforms the weakly-supervised method of~\cite{ICCV_saleh} and a DeepLab semantic segmentation network trained on synthetic data. In fact, on this dataset, it event outperforms some of the fully supervised methods that rely on annotated CamVid images for training.

\begin{table}[t]
\renewcommand{\arraystretch}{1.4}
\small
\centering
\caption{{\bf Comparison to domain adaptation and weakly-supervised methods.} All methods were trained on GTA5, except for~\protect\cite{ICCV_saleh} which does not use synthetic images and Ours which uses GTA5 for background classes and VEIS for foreground. The domain adaptation methods and Ours+Pseudo-GT make use of unlabeled real images during training. The results are reported on the Cityscapes validation set. Note that all the models below use the same backbone architecture as us (DeepLab or FCN8).
}
\label{tab:mask_eval}
\tiny
% \qquad  % get some separation between the two tabulars
 \setlength\tabcolsep{1.5pt} % default value: 6pt
 \scalebox{.95}
 {
\begin{tabular}{l|c c c c c c c c c c c c c c c c c c c |c }
 Methods & \rotatebox[origin=l]{90}{road} &\rotatebox[origin=l]{90}{side.} & \rotatebox[origin=l]{90}{buil.} & \rotatebox[origin=l]{90}{wall} & \rotatebox[origin=l]{90}{fence} & \rotatebox[origin=l]{90}{pole} & \rotatebox[origin=l]{90}{light} & \rotatebox[origin=l]{90}{sign} & \rotatebox[origin=l]{90}{Vege.} & \rotatebox[origin=l]{90}{terr.} & \rotatebox[origin=l]{90}{sky} & \rotatebox[origin=l]{90}{person} & \rotatebox[origin=l]{90}{rider} & \rotatebox[origin=l]{90}{car} & \rotatebox[origin=l]{90}{truck} & \rotatebox[origin=l]{90}{bus} & \rotatebox[origin=l]{90}{train} & \rotatebox[origin=l]{90}{motor} & \rotatebox[origin=l]{90}{bike} & \rotatebox[origin=l]{90}{mIOU} \\
\hline
Fully Sup.&   95.8  & 70.4 &  85.4 &   42.7 &   41.0 &   21.2 &   33.7 &   44.8 &   86.2 &   51.4 &   88.4 &   58.1 &   30.1 &   86.4 & 43.8 &   56.7 &   42.8 &  33.9 &  54.8 & 56.2\\
\hline
Fully Sup. Ours & 95.6 & 70.1 & 86.1 & 43.8 & 41.4 & 16.6 & 31.3 & 43.3 & 85.9 & 52.0 & 89.6 & 67.0 & 29.9 & 87.7 & 61.8 & 72.7 & 53.1 & 50.8 & 60.5 & 60.0 \\
\hline
Weakly-Sup.~\cite{ICCV_saleh}&75.9 & 1.5 & 41.7 & 14.1 & 15.3 & 6.3 & 4.4 & 7.7 & 58.4 & 12.6 & 56.2 & 16.2 & 6.1 & 41.2 & 22.7 & 16.6 & 20.4 & 15.7 & 14.9 & 23.6\\
\hline
FCNs in Wld ~\cite{wild} &70.4 & 32.4 & 62.1 & 14.9 & 5.4 & 10.9 & 14.2& 2.7 & 79.2 & 21.3 & 64.6 & 44.1 & 4.2 & 70.4 & 8.0&  7.3  & 0.0 & 3.5 & 0.0 & 27.1\\
Curriculum ~\cite{curriculum}& 74.8  & 22.0 & 71.7 & 6.0 & 11.9 & 8.4 & 16.3 & 11.1 & 75.7 & 13.3 & 66.5 & 38.0 & 9.3 & 55.2 & 18.8 & 18.9& 0.0 & 16.8 & 14.6 & 28.9 \\
ROAD ~\cite{ROAD} & 85.4 & 31.2 & 78.6 & 27.9 & 22.2  &21.9 &23.7& 11.4 &80.7 &29.3 &68.9 &48.5 &14.1 &78.0 &19.1 &23.8 &9.4 &8.3 &0.0& 35.9\\
CYCADA~\cite{cycada}&85.2 &37.2 &76.5 &21.8 &15.0 &23.8& 22.9& 21.5& 80.5& 31.3& 60.7& 50.5& 9.0& 76.9& 17.1& 28.2& 4.5 &9.8 &0.0 &35.4\\
\hline
Ours & 71.9 & 23.8 & 75.5 & 23.4 & 14.9 & 9.3 & 26.7 & 42.5 & 80.1 & 34.0 & 76.3 & 52.2 & 28.5 & 76.2 & 19.6 & 31.6 & 6.9 & 18.1 & 9.8 & 38.0 \\
Ours+Pseudo-GT& 79.8 & 29.3 & 77.8 &, 24.2 & 21.6 & 6.9 & 23.5 & 44.2 &  80.5 & 38.0 & 76.2 & 52.7 & 22.2 & 83.0 & 32.3 & 41.3 &  27.0 &  19.3 & 27.7 & 42.5  \\

\end{tabular}
}

\end{table}

\begin{table}[t]
\renewcommand{\arraystretch}{1.2}
\centering
\tiny
\caption{Comparison with fully- and weakly-supervised methods on CamVid.}
\setlength\tabcolsep{1.9pt}
\label{tab:camvid}
\scalebox{1.1}
{
\begin{tabular}{l | c @{    }c@{    } c@{    } c@{    } c@{    } c@{    } c@{    } c@{    } c@{    } c@{    } c@{    }  | c } 
\tiny
 Methods & \rotatebox[origin=l]{90}{\tiny{build.}} & \rotatebox[origin=l]{90}{\tiny{vege.}} & \rotatebox[origin=l]{90}{\tiny{sky}} & \rotatebox[origin=l]{90}{\tiny{car}} & \rotatebox[origin=l]{90}{\tiny{sign}} & \rotatebox[origin=l]{90}{\tiny{road}} & \rotatebox[origin=l]{90}{\tiny{ped.}} & \rotatebox[origin=l]{90}{\tiny{fence}} & \rotatebox[origin=l]{90}{\tiny{pole}} & \rotatebox[origin=l]{90}{\tiny{side.}} & \rotatebox[origin=l]{90}{\tiny{cyclist}} &  \rotatebox[origin=l]{0}{\tiny{mIOU}}\\
 \hline
SegNet~\cite{segnet} &  68.7 & 52.0 & 87.0 & 58.5 & 13.4 & 86.2 & 25.3 & 17.9 & 16.0 & 60.5 & 24.8  & 46.4\\
Liu and He~\cite{activeinference} &  66.8 & 66.6 & 90.1 & 62.9 & 21.4 & 85.8 & 28.0 &17.8 & 8.3 & 63.5 & 8.5 & 47.2 \\
FCN 8~\cite{FCN} &\multicolumn{10}{c}{n/a}& & 52.0\\
DeepLab-LargeFOV~\cite{deeplab,dilation}&81.5& 74.6& 89.0 &82.2 &42.3& 92.2& 48.4& 27.2& 14.3& 75.4& 50.1& 61.6\\
Dilation8~\cite{dilation}&82.6 &76.2 &89.9 &84.0 &46.9 &92.2 &56.3 &35.8 &23.4 &75.3 &55.5 &65.3\\
\hline
Weakly Sup.~\cite{ICCV_saleh} & 58.9 &   46.4 & 83.8 &   26.5 &   12.0 &   64.4 &    8.0 &   11.3 & 3.1 &   1.1  & 11.0 & 29.7\\
\hline
GTA5 & 66.6 & 53.9 & 61.4 & 70.4 & 32.8 & 80.9 & 28.2 & 24.4 & 14.6 & 57.1 & 0.0 & 44.6\\
GTA5+VEIS & 73.6 & 54.2 & 77.9 & 66.2 & 33.6 & 77.3 & 26.1 & 16.0 & 3.3 & 48.4 & 11.9 & 44.4\\
Ours & 66.3 &   55.0 &  61.9 &   73.4 &   37.4 &   82.7 &   41.4 &   23.9 &    9.2 &   57.7 &   14.9 & 47.6\\
Ours+Pseudo-GT & 72.3 & 55.2 & 72.6 & 73.1 & 37.4 & 83.9 & 39.9 & 33.2 & 1.2 & 55.5 & 12.8 & 48.8\\
\hline
\end{tabular}
}
\end{table}

\subsection{Shape vs. Texture in the Presence of Domain Shift}
In addition to Fig. 1, we experimentally show that shape is more representative than texture for foreground classes when dealing with the domain shift. To this end, first, we trained a binary VGG-16 classifier to determine whether a silhouette of a foreground object comes from real or synthetic data. We used synthetic data from our VEIS dataset and real data from Cityscapes. We found that such a classifier was unreliable to distinguish these two classes, achieving an accuracy of 70.0\% despite our best effort to train it. Note that this is better than chance because the synthetic silhouettes are perfect whereas the real ones were obtained manually. We performed the same experiment with textured foreground objects (but no background) and found that the same classifier was then successful, with an accuracy of 95.1\%. This shows that texture is indeed much more indicative of the data domain than shape, and thus supports our claim. 
As a second experiment, we trained a multi-class classifier on silhouettes of synthetic foreground VEIS objects and tested it on silhouettes of real Cityscapes objects. The resulting classifier achieved an accuracy of 81.0\% on the real data, vs 89.2\% on a validation set of synthetic samples. Training the same classifier on textured silhouettes yielded an accuracy of 83.7\% on real data and 94.2\% on synthetic data. In other words, there is a larger accuracy gap between the real and synthetic domains when training on textured data, thus further showing that shape is more robust to the domain shift.

\section{Conclusion}
We have introduced an approach to effectively leveraging synthetic training data for semantic segmentation in urban scenes, by handling the foreground classes in a detection-based manner. 
Our experiments have demonstrated that this outperforms training a standard semantic segmentation network from synthetic data and state-of-the-art domain adaptation techniques. Nevertheless, our approach is orthogonal to domain adaptation. As such, in the future we will investigate how domain adaptation can be incorporated into our framework.

\bibliographystyle{splncs04}
\bibliography{egbib}
\end{document}